\documentclass[runningheads]{llncs}
\usepackage[T1]{fontenc}
\usepackage{comment}
\usepackage{algorithm}
\usepackage{algpseudocode}
\usepackage{amsmath}
\usepackage{tabularx} 

\usepackage{graphicx}
\usepackage{listings}
\begin{document}
%
\title{PerspAct: Enhancing LLM Situated Collaboration Skills through Perspective Taking and Active Vision}
\titlerunning{PerspAct}
\author{Sabrina Patania \inst{1}, Luca Annese \inst{1}, Anita Pellegrini \inst{1},  Silvia Serino \inst{1}, Anna Lambiase \inst{1}, Luca Pallonetto \inst{2}, Silvia Rossi \inst{2}, Simone Colombani \inst{3}, Tom Foulsham \inst{4}, Azzurra Ruggeri  \inst{5} \and Dimitri Ognibene \inst{1}}
\institute{University of Milan-Bicocca, Milan, Italy\\ 
\email{\{sabrina.patania,dimitri.ognibene\}@unimb.it,}  \\
\and University of Naples Federico II, Naples, Italy \and Oversonic Robotics, Carate Brianza, Italy \and
University of Essex, Colchester, United Kingdom 
\and TUM School of Social Sciences and Technology, Munich, Germany}
%
%
%
%
\maketitle              
\begin{abstract}
Recent advances in Large Language Models (LLMs) and multimodal foundation models have significantly broadened their application in robotics and collaborative systems. However, effective multi-agent interaction necessitates robust perspective-taking capabilities, enabling models to interpret both physical and epistemic viewpoints. Current training paradigms often neglect these interactive contexts, resulting in challenges when models must reason about the subjectivity of individual perspectives or navigate environments with multiple observers. This study evaluates whether explicitly incorporating diverse points of view using the ReAct framework, an approach that integrates reasoning and acting, can enhance an LLM’s ability to understand and ground the demands of other agents. We extend the classic Director task by introducing active visual exploration across a suite of seven scenarios of increasing perspective-taking complexity. These scenarios are designed to challenge the agent’s capacity to resolve referential ambiguity based on visual access and interaction, under varying state representations and prompting strategies, including ReAct-style reasoning. Our results demonstrate that explicit perspective cues, combined with active exploration strategies, significantly improve the model’s interpretative accuracy and collaborative effectiveness. These findings highlight the potential of integrating active perception with perspective-taking mechanisms in advancing LLMs’ application in robotics and multi-agent systems, setting a foundation for future research into adaptive and context-aware AI systems.

\keywords{perspective taking  \and LLMs \and active vision \and theory of mind.}
\end{abstract}
\section{Introduction}

Effective interaction in multi-agent systems, particularly those involving human-AI collaboration or autonomous robotic agents, demands more than basic task execution capabilities. Such interactions often require the artificial agent to exhibit a nuanced understanding of the perspectives of other agents within the environment.  In this context, perspective-taking refers to an agent’s ability to interpret and adopt both the physical and epistemic viewpoints of others, which is crucial for seamless and contextually appropriate responses. Our study explores how integrating perspective-taking mechanisms within a dynamic reasoning framework can enhance an agent’s capacity to operate effectively in complex, interactive scenarios.

To address this challenge, we build upon the ReAct (Reason and Act) framework \cite{yao2023react}, which offers a structured approach for integrating cognitive reasoning with environmental actions. Unlike traditional models that separate perception from action, the ReAct framework allows agents to alternate between interpreting instructions and engaging with their surroundings to gather additional information. This dual capability is particularly valuable in scenarios where an agent must infer another's perspective, as it can both simulate different viewpoints cognitively and physically explore the environment to align its perception with that of another agent.

Our study introduces a modified version of the classic director task \cite{keysar2000taking}, adapting it to more ecological and realistic scenarios \cite{sarthou2021director}. The classical paradigm investigates individuals' ability to consider others' visual perspectives during communication. Participants follow instructions from a "director" to move objects within a grid, where some objects are occluded from the director's view. Critical trials require participants to select appropriate objects considering the director’s point of view (e.g., selecting a medium-sized candle when instructed to move "the small candle", as the actually smallest candle is hidden from the director's perspective) \cite{keysar2000taking}\cite{keysar2003limits}.

Our experimental setup involves two agents, director and matcher. The former is tasked with guiding the matcher to retrieve a specific object within a shared environment. However, the environment is partially observable, meaning that the matcher may not initially see all relevant objects or may have a different spatial understanding compared to the director. Success in this task depends on effective communication, where the director must provide instructions that account for the matcher’s perspective, and the matcher must interpret these instructions accurately.

To further enhance this interaction, a distinguishing feature of our approach is the incorporation of active visual exploration. Rather than passively receiving and processing static environmental inputs, the matcher agent can dynamically adjust its viewpoint to better understand the scene from the director’s perspective. This ability to move within the environment — whether by virtually panning a camera, shifting position, or focusing on specific objects — enables the agent to bridge gaps in its understanding and make more informed decisions. This active vision component simulates a human-like strategy of physically adjusting one’s position to gain a clearer view of a situation, which is particularly useful in scenarios where visual occlusion or complex spatial arrangements are involved.

Another feature of our approach is a sophisticated communication strategy that goes beyond simple command-response interactions. When ambiguity arises, the matcher can initiate clarification dialogues, engaging the director in a process that reflects a rudimentary form of Theory of Mind (ToM), the ability to represent, understand, and reason about others’ mental states \cite{frith2005theory}. This process involves considering what the director might know, believe, or intend, and not only interpreting the explicit content of the director's instructions. By embedding perspective-taking directly into both the reasoning and action phases, our model aims to improve not only the efficiency of task completion but also the quality and naturalness of interactions between agents, thereby enhancing collaboration in complex environments .

The goal of this research is to show that combining perspective-taking with active exploration strategies within the ReAct framework can improve an agent’s adaptability and collaborative effectiveness. This could lead to more adaptable, socially aware robotic systems, that can navigate complex environments and better understand human communication.

\section{Related Work}

In recent years, there has been growing interest in the application of LLMs and multimodal foundation models in robotics, and of collaborative systems for high-level reasoning, perception, and decision-making \cite{ognibene2025scoopframeworkproactivecollaboration}. These models are pre-trained on vast amounts of internet-scale data, and exhibit impressive generalization capabilities \cite{brown2020language}, enabling robots to handle a wide range of open-ended scenarios. Models such as SayCan \cite{ahn2022can} and Inner Monologue \cite{huang2022inner} demonstrate how LLMs can break down abstract goals into practical steps by combining high-level reasoning with grounded robotic actions.

A core component of effective multi-agent interaction is perspective-taking, namely, the ability to represent a situation from an alternate viewpoint \cite{grice1975logic}. It includes two different but related processes: visual perspective-taking, with a distinction between Level-1 (determining what objects are visible to others) and Level-2 processes (representing how visible objects appear from another's viewpoint); and spatial perspective-taking, which involves representing relative spatial relationships between agents and objects, typically represented through egocentric and allocentric reference frames \cite{flavell1981young}. Visual perspective-taking has particularly been linked to  ToM. Specifically, Level-2 perspective-taking is related to ToM, because both functions require a decoupled representation from one's perspective or belief. Within frameworks such as ReAct \cite{yao2023react}, perspective-taking is conceptualized as a specialized form of reasoning that works alongside acting. This approach allows models to continuously update their knowledge states through real-time interactions with their environment.

Efforts to enhance perspective-taking in LLMs have largely focused on language-based tasks. Early studies using false-belief tasks reveal that while older models (e.g. GPT-2, early GPT-3) performed poorly, the latest systems (e.g. GPT-4) show emergent - though unstable - ToM capabilities \cite{kosinski2024evaluating}. Techniques such as the SimToM prompting framework \cite{wilf2023think} have further improved performance by explicitly instructing models to simulate a character’s perspective, mitigating the injection of omniscient background knowledge by the system itself.

Parallel research in vision-language models (VLMs) has explored spatial perspective-taking. Datasets like Isle-Bricks and Isle-Dots \cite{goral2024seeing} demonstrate that, although many VLMs can recognize objects within a scene, they often struggle with inferring what an observer can or cannot see. Advanced systems like GPT-4V excel on simple Level-1 tasks but experience significant performance drops on more complex Level-2 tasks involving mental rotation and continuous viewpoint changes \cite{leonard2024failures}.

Beyond static analysis, active visual exploration represents an added layer of complexity. Benchmarks such as ActiView \cite{wang2024actiview} require models to actively adjust their perceptual focus (e.g., by zooming in, or by shifting their view), to gather additional information. Although some models show an advantage when processing sequential observations, overall performance in active visual exploration remains far below human capabilities.

In addressing these challenges, recent work has explored strategies to resolve ambiguity in multi-agent settings. For example, prior studies have investigated using LLMs as active Bayesian filters for information acquisition and integration \cite{patania2024large}. Such a mechanism not only addresses ambiguous or incomplete perceptual inputs but also naturally complements robust perspective-taking, ultimately enhancing multi-agent interaction in dynamic environments.

These studies illustrate both the potential and current limitations of LLMs and VLMs in modelling perspective-taking and active exploration. While advances in prompting and dynamic reasoning have improved performance on discrete tasks, significant challenges remain in achieving a robust, context-sensitive understanding of multiple viewpoints in dynamic, real-world environments.

\section{Method}

\subsection{Task}
Our research utilizes an extended version of the classic director's task, reformulated to investigate perspective-taking capabilities within a more ecologically valid and complex environment. This adaptation enhances the traditional paradigm by incorporating elements of active exploration, partial observability, and dynamic communication between agents.

\subsection{Experimental Environment}

To investigate perspective-taking within goal-directed interaction, we developed a simulated household environment using the Planning Domain Definition Language (PDDL) \cite{aeronautiques1998pddl}. The environment models a shared space occupied by two agents: a Director, who possesses knowledge of both the identity and location of a target object, and a Matcher, who must retrieve the object using limited perceptual information and dialogue. The setting comprises multiple spaces containing various objects and containers, some of which obscure visibility—requiring the Matcher to engage in active exploration and adopt the Director’s perspective.

The PDDL domain captures both the spatial layout and the asymmetry in perceptual access. Each agent can observe the contents of its current location as well as adjacent areas, creating a partially overlapping field of view. This limited shared perspective enables basic grounding while still necessitating inferential reasoning about what the other agent can or cannot perceive.

Ambiguity is commonly introduced through the inclusion of duplicate object types (e.g., two ties differing in colour), requiring the Matcher to resolve referential uncertainty through dialogue or strategic exploration—beyond mere spatial proximity or visual cues.

To systematically vary the demands on perspective-taking, we defined seven task types that differ in referential ambiguity, the necessity to reason about another agent’s viewpoint, and the proximity of candidate objects:
\begin{description}      
    \item[Base] Both agents can see both objects, and the Matcher is near the correct one. The Director explicitly identifies the intended object. This scenario tests the Matcher’s ability to resolve ambiguity through linguistic disambiguation rather than spatial reasoning alone.       
    \item[Persp] The Matcher can see both candidate objects, while the Director sees only one. This situation requires the Matcher to select the object that is visible to both agents, demonstrating a basic form of shared attention inference.
    \item[Distractor] The Matcher sees an incorrect item, while the Director sees only the correct one, located elsewhere. This case requires the Matcher to recognize that its own perception may be insufficient and be open to correction via the Director’s response.
    \item[Far] Both the Matcher and Director have visual access to both objects; however, the Matcher is positioned closer to the incorrect item. \textit{In this ambiguous setup, optimal behavior involves asking for clarification rather than assuming proximity indicates relevance.}
    \item[Near] Similar to \textbf{Far}, both agents see both objects, but this time the Matcher is closer to the correct one. Rather than assuming, the Matcher should ideally seek confirmation, showing an awareness of potential mismatches in intent.
    \item[Hidden] The object is entirely outside the Matcher’s field of view but visible to the Director. The Matcher must actively explore the environment or ask the director to locate the object, relying on movement and perspective-taking rather than direct perception.
    \item[NotThat] The Matcher can only see an incorrect object, while the Director sees both the correct and incorrect items. The Matcher, unaware of the alternative, is likely to make an incorrect selection and must adjust behavior upon receiving feedback.
\end{description}

The Matcher begins each trial at a certain location and must perform a sequence of actions: moving, opening containers, and optionally asking questions, to infer the correct object. This design enables controlled evaluation of perspective-taking under increasing complexity, balancing grounded environment interaction with higher-order inference demands.
\subsection{Agents}
We employed two different agents based on the DSPy \cite{khattab2023dspy} framework to simulate the Matcher. The first is a baseline DSPy agent, which operates purely as a zero-shot language model. It generates responses directly from a prompt that encodes the current instruction and environment state, without any intermediate reasoning or memory of previous steps. The second agent is built using the DSPy ReAct module, which extends this approach by alternating between natural-language "thoughts" and grounded "actions." While it also operates in a zero-shot fashion, the ReAct agent is guided by a declarative program that structures the reasoning process, prompting the model to reflect, take an action, observe the result, and iterate. Importantly, neither agent is pre-trained or fine-tuned on examples; instead, they rely on declarative instructions and structured prompting at inference time.
The ReAct framework addresses limitations in standard language model prompting techniques by integrating reasoning with action-taking capabilities. In traditional approaches such as chain-of-thought prompting, language models generate reasoning traces but cannot access external information during their reasoning process, often leading to hallucinations or reliance on outdated knowledge.

The third agent is the Director, implemented as a stateless OpenAI language model. Unlike the Matcher agents, the Director has no capacity to move or interact physically with the environment. Its sole role is to respond to clarification queries posed by the Matcher. It operates using a fixed, carefully engineered prompt, tailored to ensure consistent and accurate responses grounded in the current environment configuration. This prompt encodes the Director's knowledge of the task and objects visibility, making it a stable information source during ambiguous scenarios.

\section{Experiments}

\begin{figure}[ht]
    \centering
    \includegraphics[width=1\textwidth]{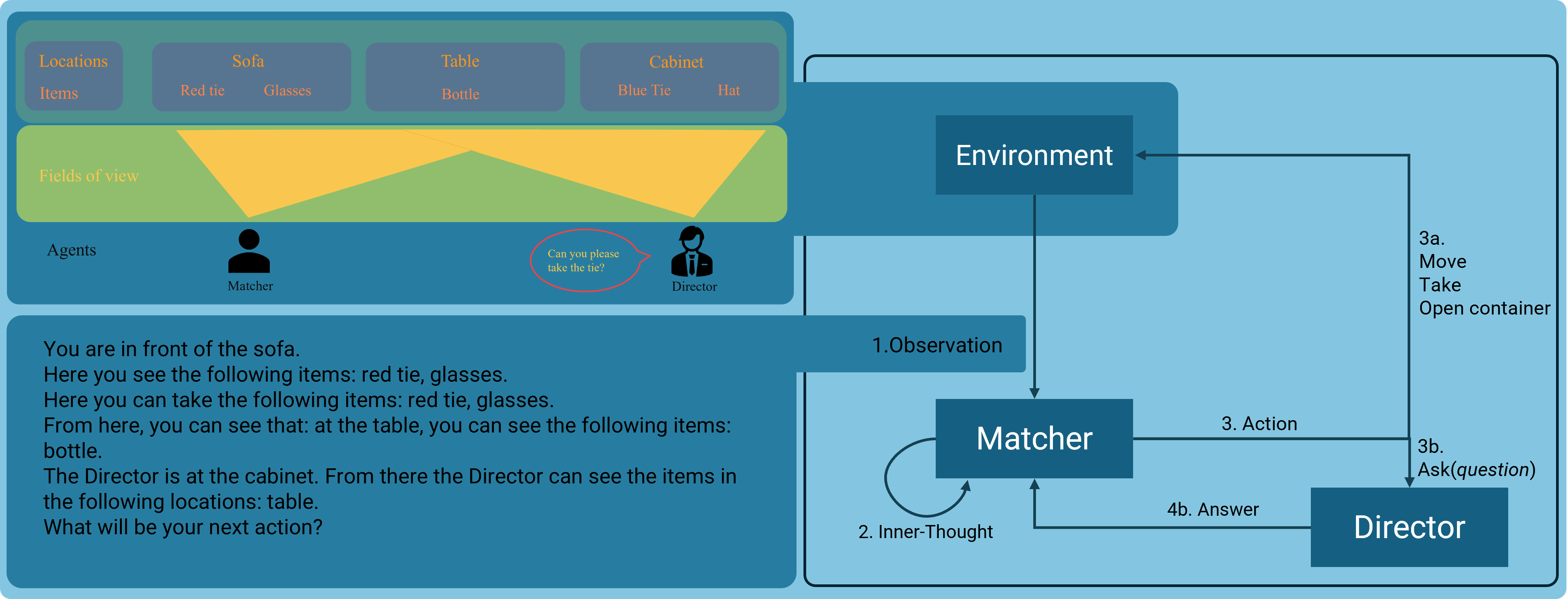}
    \caption{Schematic view of the experiment. The top-left panel illustrates a sample environment highlighting the perspective of both the Matcher and the Director (yellow cones). The bottom-left panel presents a sample of an observation the Environment produces sequentially. The observation is provided to the Matcher in a natural language fashion, via a fixed template that ri-elaborate the output of the PDDL-defined environment. Finally, the right-panel, shows the action-perception cycle of the Matcher: the Matcher firstly receive observe the Environment and then \textit{think about} the possible course of actions. After that, the Matcher decides whether questioning the Director for clues of engage with the environment. The \textit{inner-thought} cycle depicted here, refers to the ReAct-based Matcher, that each time reflects before taking the next action.}
    \label{fig:example}
\end{figure}
We designed our experiments to evaluate the capacity of large language models to perform perspective-taking in an interactive setting, and to assess the extent to which this ability emerges from situated reasoning alone. In our referential household environment, an LLM-based agent (the Matcher) must interpret underspecified natural-language instructions issued by a stationary Director agent, which cannot move but can answer clarification queries. The Matcher must rely solely on its own observations and interactions with the Director to resolve ambiguity. By varying task difficulty and the level of ambiguity, we analysed how well the Matcher can infer the Director’s perspective and act accordingly.

\lstset{escapeinside={(*@}{@*)}}
\begin{lstlisting}[language=,
  caption={Matcher's Prompt Regarding the Request from the Director},
  label={lst:prompt_topdown}]
You are the Matcher in this task. 
The task is to take an object that the Director asks you about.
The object the Director is referring to is the (*@\textit{object}@*). 
The Director cannot act in the environment and only knows 
what he sees from his location.
\end{lstlisting}
To evaluate the LLMs behavior, we conducted a series of multi-step simulations (5 trials per each scenario, with up to 15 time steps) using two agent configurations: PerspAct, the ReAct-based agent and the baseline agent, both based on the GPT o3-mini model and using the same prompts.

To evaluate the Matcher’s performance across different perspective-taking demands, we employed three primary metrics. First, we measured the percentage of trials where the agent’s first attempt involved taking the wrong object (Table \ref{tab1_fail}), as an indicator of initial disambiguation success. Second, we tracked the average number of steps required to complete each task, capturing overall efficiency (Table \ref{tab1_step}). Lastly, we counted the average number of ask-actions, reflecting the agent’s tendency to seek clarification from the Director (Table \ref{tab1_ask}). Together, these metrics allow us to assess not only task success, but also the agent's reasoning strategy, balancing initiative, caution, and information gathering, under varying visibility and ambiguity conditions.
\begin{table}[H]
\caption{First Take on Incorrect Target (\%) across Scenarios}
\label{tab1_fail}
\begin{tabularx}{\textwidth}{|p{1.6cm}|X|X|X|X|X|X|X|}
\hline
\textbf{Agent} & \textbf{Base} & \textbf{Persp} & \textbf{Dist} & \textbf{Far} & \textbf{Near} & \textbf{Hidd} & \textbf{NotThat} \\
\hline
Base &  0 & 0 & 100 & 100 & 0 & 0 & 100 \\
\hline
PerspAct &  0 & 0 & 60 & 100 & 0 & 0 & 100 \\
\hline
\end{tabularx}
\end{table}
\begin{table}[H]
\caption{Average Number of Steps}
\label{tab1_step}
\begin{tabularx}{\textwidth}{|p{1.6cm}|X|X|X|X|X|X|X|}
\hline
\textbf{Agent} & \textbf{Base} & \textbf{Persp} & \textbf{Dist} & \textbf{Far} & \textbf{Near} & \textbf{Hidd} & \textbf{NotThat} \\
\hline
Base &  3 & 2 & 6.4 & 5.6 & 2 & 4 & 6.6 \\
\hline
PerspAct &  3 & 2 & 5.6 & 5.4 & 2 & 4 & 6.8 \\
\hline
\end{tabularx}
\end{table}
\begin{table}[H]
\caption{Average Number of \textit{Ask} Actions}
\label{tab1_ask}
\begin{tabularx}{\textwidth}{|p{1.6cm}|X|X|X|X|X|X|X|}
\hline
\textbf{Agent} & \textbf{Base} & \textbf{Persp} & \textbf{Dist} & \textbf{Far} & \textbf{Near} & \textbf{Hidd} & \textbf{NotThat} \\
\hline
Base &  1 & 1 & 3.4 & 2.6 & 1 & 2 & 2.6 \\
\hline
PerspAct &  1 & 1 & 3 & 2.6 & 1 & 2 & 2.8 \\
\hline
\end{tabularx}
\end{table}
To establish a reference baseline, we employed the Fast Downward planner configured with A* search and the FF heuristic to compute optimal action sequences. The planning problems were adapted, without changing the overall environment structure, to allow the computation of optimal perspective-taking plans for each scenario. In doing so, the PDDL domain was specifically extended to accommodate epistemic behaviors: in scenarios where perspective simulation alone could not resolve ambiguity and MOVE actions were inadequate for disambiguation, the planner was permitted to use ASK actions. These optimal plans, guided by perspective-taking, served as expert baselines for grounded reasoning. These represent the most efficient sequences that a theoretically omniscient agent would follow to resolve ambiguities and correctly retrieve the target object.
\begin{table}[H]
\caption{Optimal Perspective Taking Behavior}
\label{tab4_plann}
\begin{tabularx}{\textwidth}{|p{1.5cm}|X|X|X|X|X|X|X|}
\hline\textbf{Agent} & \textbf{Base} & \textbf{Persp} & \textbf{Dist} & \textbf{Far} & \textbf{Near} & \textbf{Hidd} & \textbf{NotThat}
 \\
\hline
\#Steps &  1 & 2 & 2 & 3 & 2 & 2 & 2\\
\hline
\#Ask &  0 & 1 & 0 & 1 & 0 & 0 & 1  \\
\hline
\#Move &  0 & 0 & 1 & 1 & 1 & 1 & 0 \\
\hline
\end{tabularx}
\end{table}
Both Base and PerspAct agents perform flawlessly in conditions where perceptual access alone suffices (Base, Persp, Near, Hidd), confirming a baseline level of perspective-taking. However, in Distractor and NotThat, which require deeper perspective reasoning, both agents fail to reliably disambiguate targets. The PerspAct agent shows modest gains in Distractor (60\% vs. 100\% incorrect first picks), but neither reliably infers beyond the visible scene.

Efficiency-wise, PerspAct completes tasks with slightly fewer steps and Ask actions in complex scenarios, suggesting more purposeful interactions. Still, both fall short compared to the optimal plans computed by the Fast Downward planner (Table~\ref{tab4_plann}), which resolve ambiguity with minimal steps and strategic Ask/Move actions. The main difference appears to be in the number of questions necessary to complete the task, showing the necessity for more effective querying strategies~\cite{patania2024large,bertolazzi-etal-2023-chatgpts}.

This gap highlights that while agents exhibit reactive perspective sensitivity, they lack the capacity for goal-directed, epistemic behavior as demonstrated by the optimal plans. Their failure to infer unseen referents or strategically query the Director in ambiguous contexts limits their perspective-taking to surface-level heuristics.

\section{Discussion}
Across all tested configurations, both PerspAct, the ReAct-based agent, and the classic LLM agent exhibit broadly similar behavior with respect to perspective-taking. In tasks where perceptual cues alone are sufficient, such as Persp, Hidden, Base, and Near, both agents achieve perfect accuracy using minimal actions, suggesting that basic forms of perspective-awareness are accessible to the LLMs even without explicit perspective-taking manipulation or training. This implies that surface-level perspective-taking strategies (i.e., reasoning based on one’s own field of view and interpreting straightforward instructions) are within the reach of current models, even without explicit demonstrations.

However, in the more challenging scenarios involving conflicting visibility, such as NotThat and Distractor, both agents struggle to disambiguate and acquire missing information effectively. These tasks require reasoning beyond the current perceptual field (i.e., inferring that the object mentioned by the Director might not be directly visible). In these cases, agents display a conservative bias toward immediately visible candidate objects, failing to account for unseen alternatives. This leads to persistent errors that reveal a lack of robust perspective inference. Importantly, while task completion is eventually achieved in all cases, the reasoning remains shallow and reactive.

When comparing agents, the ReAct-based Matcher generally completes complex scenarios more efficiently, requiring fewer steps and clarification queries, suggesting that interleaving reasoning with interaction can support better procedural decision-making. Yet, the fundamental limitations in perspective-taking persist across both agents: neither is able to reliably infer the presence of a non-visible target when a similar visible item is present, often defaulting to the reachable object instead.
In the NotThat condition, such behavior may still be considered rational: assuming the Director meant the most salient, reachable referent may be a cost-efficient heuristic when additional possibilities are epistemically inaccessible. In contrast, the Distractor condition arguably reflects a joint failure of both agents: the Matcher appears to overestimate the Director’s contextual knowledge, while the Director fails to tailor the instruction to account for the Matcher’s limited perception.



Despite these promising results, this study represents only a preliminary step towards the integration of perspective-taking in language models. Several limitations and potential directions for future research must be considered. One immediate avenue for improvement involves incorporating learning mechanisms within the prompting process. Methods such as in-context learning could enable models to refine their performance over successive interactions by being exposed to high-quality examples that illustrate effective perspective-taking strategies. By structuring prompts with curated demonstrations of successful reasoning and exploration, future models could further enhance their interpretative accuracy and collaborative performance.


The ecological validity of our experimental setup could also be improved by expanding to more challenging and realistic environments. While our current task formulation within a household simulation captures key aspects of real-world interactions, future iterations could incorporate larger, more dynamic environments with increased uncertainty. Additionally, the action space could be extended to include a broader set of exploration and interaction capabilities, further testing the adaptability of the framework under complex and ambiguous conditions.

Finally, a critical next step involves transitioning from simulated environments to real-world robotic implementations. While the current study was conducted in a structured, controlled digital setting, applying the same methodology to physical robots would introduce new challenges related to perception, motor execution, and real-time reasoning constraints. By progressively bridging the gap from simulated agents to embodied AI, future research could explore how perspective-taking and interactive reasoning translate into physical actions, ultimately enabling more effective and adaptable robotic systems.

In conclusion, while the ReAct agent shows modest gains in efficiency, both systems are currently limited in their capacity to simulate full-fledged perspective-taking, particularly when it involves reasoning about non-visible alternatives or considering differences in what other agents can observe. Moreover, significant work remains in refining these techniques, expanding their applicability, and ultimately transitioning them to embodied agents capable of operating in dynamic, real-world environments. Future research should focus on enhancing learning through prompting, optimizing adaptive reasoning strategies, and bridging the gap between digital simulations and robotic implementations.

\subsubsection{\ackname}
This work has been supported by the European Union - Next Generation EU, Mission 4 Component 1, PRIN 2022 TrustPACTX CUP E53D23007850001 and by Volkswagen Foundation grant "Developing an Artificial Social Childhood (ASC) to improve AI causal reasoning, information gathering and decision making” in the funding program ‘Open Up – New Research Spaces for the Humanities and Cultural Studies’ ref 9E530.


%
%
%
\bibliographystyle{splncs04}
\bibliography{mybibliography}
\end{document}